\documentclass[10pt,twocolumn,letterpaper]{article}

\usepackage{cvpr}
\usepackage{times}
\usepackage{epsfig}
\usepackage{graphicx}
\usepackage{amsmath}
\usepackage{amssymb}

\usepackage{multirow}
\usepackage{bbding}
\usepackage{pifont}
\usepackage{wrapfig}
\usepackage{algorithm}
\usepackage{algpseudocode}
\usepackage{tabularx, booktabs} 

\usepackage{enumitem}

\usepackage[pagebackref=true,breaklinks=true,letterpaper=true,colorlinks,bookmarks=false]{hyperref}

\cvprfinalcopy 

\def\cvprPaperID{} 

\begin{document}

\title{Towards Instance-level Image-to-Image Translation}

\author{Zhiqiang Shen$^{1,3}$\thanks{Work done during internship at SenseTime.}~, ~Mingyang Huang$^2$, ~ Jianping Shi$^{2}$, ~ Xiangyang Xue$^3$, ~ Thomas Huang$^1$ \\
		{$^1$University of Illinois at Urbana-Champaign,
			$^2$SenseTime Research,
			$^3$Fudan University}\\
		\tt\small zhiqiangshen0214@gmail.com
		\tt\small \{huangmingyang, shijianping\}@sensetime.com \\
		\tt\small xyxue@fudan.edu.cn
		\tt\small t-huang1@illinois.edu
}


\maketitle

\begin{abstract}
Unpaired Image-to-image Translation is a new rising and challenging vision problem that aims to learn a mapping between unaligned image pairs in diverse domains. Recent advances in this field like MUNIT~\cite{huang2018multimodal} and DRIT~\cite{lee2018diverse} mainly focus on disentangling content and style/attribute from a given image first, then directly adopting the global style to guide the model to synthesize new domain images. However, this kind of approaches severely incurs contradiction if the target domain images are content-rich with multiple discrepant objects. In this paper, we present a simple yet effective instance-aware image-to-image translation approach (INIT), which employs the fine-grained local (instance) and global styles to the target image spatially. The proposed INIT exhibits three import advantages: (1) the instance-level objective loss can help learn a more accurate reconstruction and incorporate diverse attributes of objects; (2) the styles used for target domain of local/global areas are from  corresponding spatial regions in source domain, which intuitively is a more reasonable mapping; (3) the joint training process can benefit both fine and coarse granularity and incorporates instance information to improve the quality of global translation. We also collect a large-scale benchmark\footnote{contains 155,529 high-resolution natural images across four different modalities with object bounding box annotations. A summary of the entire dataset is provided in the following sections. Project page: \url{http://zhiqiangshen.com/projects/INIT/index.html}.} for the new instance-level translation task. We observe that our synthetic images can even benefit real-world vision tasks like generic object detection.
\end{abstract}

\section{Introduction}
In the recent years, Image-to-Image (I2I) translation has received significant attention in computer vision community, since many vision and graphics problems can be formulated as an I2I translation problem like super-resolution, neural style transfer, colorization, etc. This technique has also been adapted to the relevant fields such as medical image processing~\cite{zhang2018translating} to further improve the medical volumes segmentation performance. In general, Pix2pix~\cite{isola2017image} is regarded as the first unified framework for I2I translation which adopts conditional generative adversarial networks~\cite{mirza2014conditional} for image generation, while it requires the paired examples during training process. A more general and challenging setting is the unpaired I2I translation, where the paired data is unavailable.

\begin{figure}[t]
	\centering
	\includegraphics[width=0.38\textwidth]{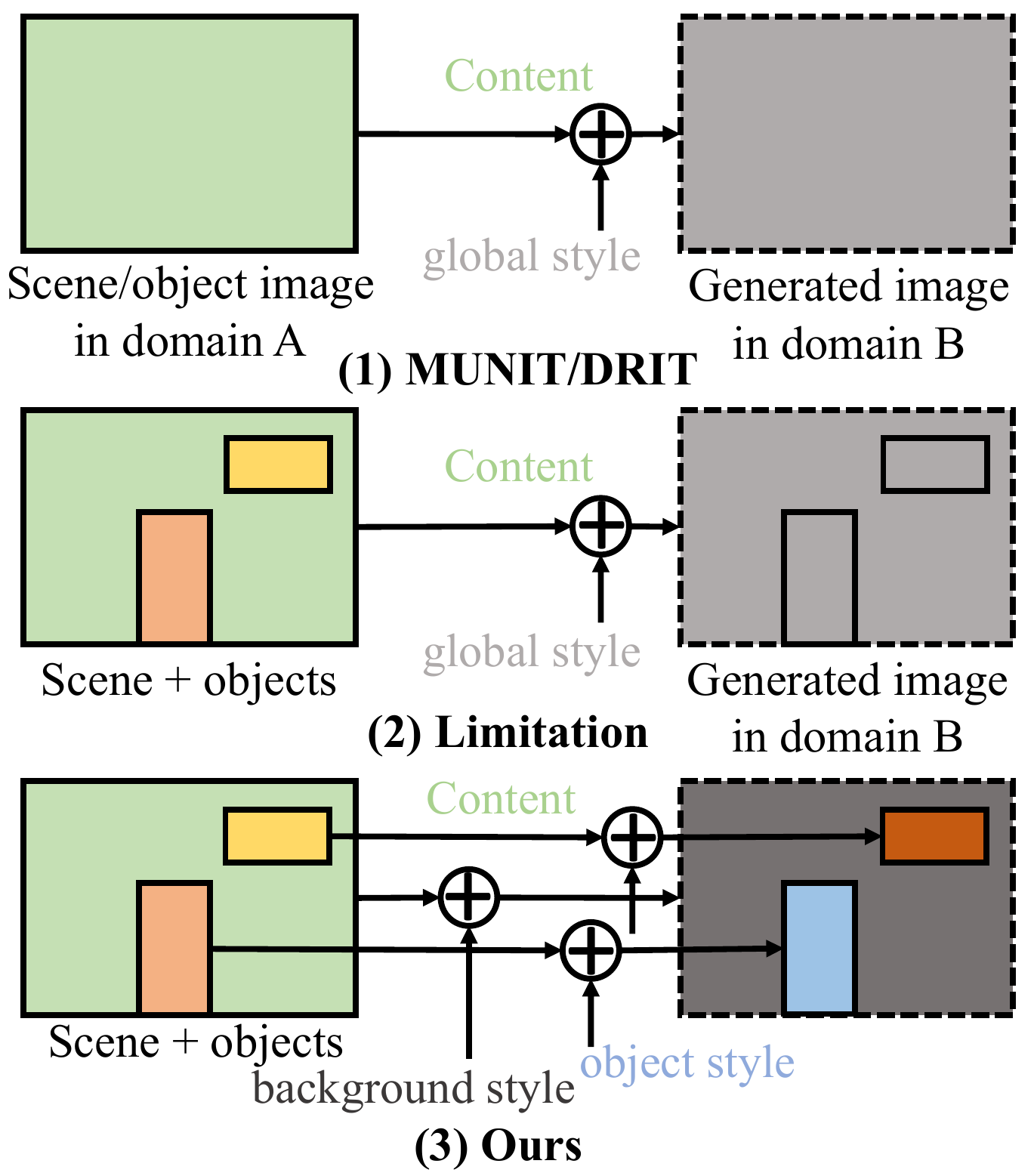}
	\vspace{-0.06in}
	\caption{Illustration of the motivation of our method. (1) MUNIT~\cite{huang2018multimodal}/DRIT~\cite{lee2018diverse} methods; (2) their limitation; and (3) our solution for instance-level translation. More details can be referred to the text.} 
	\label{limitation}
	\vspace{-0.16in}
\end{figure}

Several recent efforts~\cite{CycleGAN2017,liu2017unsupervised,huang2018multimodal,lee2018diverse,almahairi2018augmented} have been made on this direction and achieved very promising results. For instance, CycleGAN~\cite{CycleGAN2017} proposed the cycle consistency loss to enforce the learning process that if an image is translated to the target domain by learning a mapping and translated back with an inverse mapping, the output should be the original image. Furthermore, CycleGAN assumes the latent spaces are separate of the two mappings. In contrast, UNIT~\cite{liu2017unsupervised} assumes two domain images can be mapped onto a shared latent space. MUNIT~\cite{huang2018multimodal} and DRIT~\cite{lee2018diverse} further postulate that the latent spaces can be disentangled to a shared content space and a domain-specific attribute space.

However, all of these methods thus far have focused on migrating styles or attributes onto the entire images. As shown in Fig.~\ref{limitation} (1), they work well on the unified-style scenes or relatively content-simple scenarios due to the consistent pattern across various spatial areas in an image, while this is not true for the complex structure images with multiple objects since the stylistic vision disparity between objects and background in an image is always huge or even totally different, as in Fig.~\ref{limitation} (2).

\begin{figure}[t]
	\centering
	\includegraphics[width=0.40\textwidth]{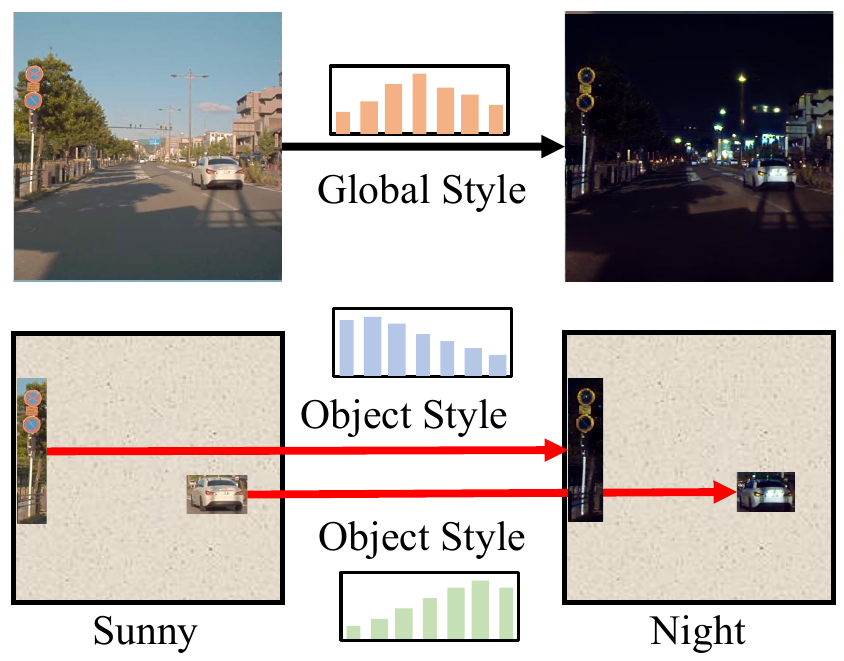}
	\vspace{-0.06in}
	\caption{A natural image example of our I2I translation.} 
	\label{motivation}
	\vspace{-0.16in}
\end{figure}

To address the aforementioned limitation, in this paper we present a method that can translate objects and background/global areas separately with different style codes as in Fig.~\ref{limitation} (3), and still training in an end-to-end manner. The motivation of our method is illustrated in Fig.~\ref{motivation}. Instead of using the global style, we use instance-level style vectors that can provide more accurate guidance for visually related object generation in target domain. We argue that styles should be diverse for different objects, background or global image, meaning that the style codes should not be identical for the entire image. More specifically, a car from ``sunny'' to the ``night'' domain should have different style codes comparing to the global image translation between these two domains. Our method achieves this goal by involving the instance-level styles. Given a  pair of unaligned images and object locations, we first apply our encoders to obtain the intermediate global and instance level content and style vectors separately. Then we utilize the cross-domain mapping to obtain the target domain images by swapping the style/attribute vectors. Our swapping strategy is introduced with more details in Sec.~\ref{method}. The main advantage of our method is the exploration and usage of object level styles, which affects and guides the generation of target domain objects directly.
Certainly, we can also apply the global style for target objects to enforce the model to learn more diverse results.

In summary, our contributions are three fold:

\vspace{-0.08in}
\begin{itemize}
	\addtolength{\itemsep}{-0.05in}
	\item  We propel I2I translation problem step forward to instance-level such that the constraints could be exploited on both instance and global-level attributes by adopting the proposed compound loss.
	\item We conduct extensive qualitative and quantitative experiments to demonstrate that our approach can surpass against the baseline I2I translation methods. Our synthetic images can be even beneficial to other vision tasks such as generic object detection, and further improve the performance.
	\item We introduce a large-scale, multimodal, highly varied I2I translation dataset, containing $\sim$155k streetscape images across four domains. Our dataset not only includes the domain category labels, but also provides the detailed object bounding box annotations, which will benefit the instance-level I2I translation problem.
\end{itemize}

\section{Related Work}

\noindent{\textbf{Image-to-Image Translation.}}
The goal of I2I translation is to learn the mapping between two different domains. Pix2pix~\cite{isola2017image} first proposes to use conditional generative adversarial networks~\cite{mirza2014conditional} to model the mapping function from input to output images. Inspired by Pix2pix, some works further adapt it to a variety of relevant tasks, such as semantic layouts $\to$ scenes~\cite{karacan2016learning}, sketches $\to$ photographs~\cite{sangkloy2017scribbler}, etc. Despite popular usage, the major weaknesses of these methods are that they require the paired training examples and the outputs are single-modal. In order to produce multimodal and more diverse images, BicycleGAN~\cite{zhu2017toward} encourages the bijective consistency between the latent and target spaces to avoid the mode collapse problem. A generator learns to map the given source image, combined with a low-dimensional latent code, to the output during training. While this method still needs the paired training data. 

\begin{table*}[h]
	\centering

	\begin{tabular}{c|c|c|c|c|c}
		\hline
		\bf Datasets & \bf Paired  & \bf Resolution &   \bf Bbox annotations &  \bf Modalities & \bf \# images  \\ \hline
		edge$ \leftrightarrow $shoes~\cite{isola2017image} 	& \Checkmark&low& -  & \{edge, shoes\}  & 50,000 \\ \hline
		edge$ \leftrightarrow $handbags~\cite{isola2017image} 	& \Checkmark&low& -  & \{edge, handbags\}  & 137,000 \\ \hline
		CMP Facades~\cite{Tylecek13}	& \Checkmark&HD& -  & \{facade, semantic map\}  & 606 \\ \hline
		Yosemite (summer$ \leftrightarrow $winter)~\cite{CycleGAN2017} 	&  \ding{55}  & HD   &   -   & \{summer, winter\} & 2,127\\ \hline 
		Yosemite$^*$ (MUNIT)~\cite{huang2018multimodal}  &   \ding{55} &  HD  & - &  \{summer, winter\}& 5,638 \\ \hline
		Cityscapes~\cite{cordts2016cityscapes}&  \Checkmark   & HD   & \Checkmark &   \{ semantic, realistic\} & 3,475 \\ \hline
		Transient Attributes~\cite{Laffont14}&  \Checkmark    &  HD  & \ding{55}  &  \{40 transient attributes\}  &  8,571 \\ \hline
		\bf Ours	&   \ding{55}    &   HD$^\dagger$ & \Checkmark   &    \{sunny, night, cloudy, rainy\}   & \bf 155,529 \\ \hline
	\end{tabular}
	\caption{{\bf Feature-by-feature comparison of popular I2I translation datasets.} Our dataset contains four relevant but visually-different  domains: sunny, night, cloudy and rainy. $^\dagger$The images in our dataset contain two types of resolutions: 1208$\times$1920 and 3000$\times$4000.}
	\label{feature}
		\vspace{-1.5ex}
\end{table*}

Recently, CycleGAN~\cite{CycleGAN2017} is proposed to tackle the unpaired I2I translation problem by using the cycle consistency loss. UNIT~\cite{liu2017unsupervised} further makes a share-latent assumption and adopts Coupled GAN in their method. To address the multimodal problem, MUNIT~\cite{huang2018multimodal}, DRIT~\cite{lee2018diverse}, Augmented CycleGAN~\cite{almahairi2018augmented}, etc. adopt a disentangled representation to further learn diverse I2I translation from unpaired training data.

\noindent{\textbf{Instance-level Image-to-Image Translation.}}
To the best of our knowledge, there are so far very few efforts on the instance-level I2I translation problem. Perhaps the most similar to our work is the recently proposed {\em Insta}GAN~\cite{mo2018instanceaware}, which utilizes the object segmentation masks to translate both an image and the corresponding set of instance attributes while maintaining the permutation invariance property of instances. A context preserving loss  is designed to encourage model to learn the identity function outside of target instances. The main difference with ours is that {\em insta}GAN cannot translate different domains for an entire image sufficiently. They focus on translating instances and maintain the outside areas, in contrast, our method can translate instances and outside areas simultaneously and make global images more realistic. Furthermore, {\em Insta}GAN is built on the CycleGAN~\cite{CycleGAN2017}, which is single modal, while
we choose to leverage the MUNIT~\cite{huang2018multimodal} and DRIT~\cite{lee2018diverse} to build our INIT, thus our method inherits  multimodal and unsupervised properties, meanwhile, produces more diverse and higher quality images.

Some other existing works~\cite{ma2018gan,li2018beautygan} are more or less related to this paper. For instance, DA-GAN~\cite{ma2018gan} learns a deep attention encoder to enable the instance-level translation, which is unable to handle the multi-instance and complex circumstance. BeautyGAN~\cite{li2018beautygan} focuses on facial makeup transfer by employing histogram loss with face parsing mask.

\begin{figure*}[t]
	\centering
	\includegraphics[width=0.98\textwidth]{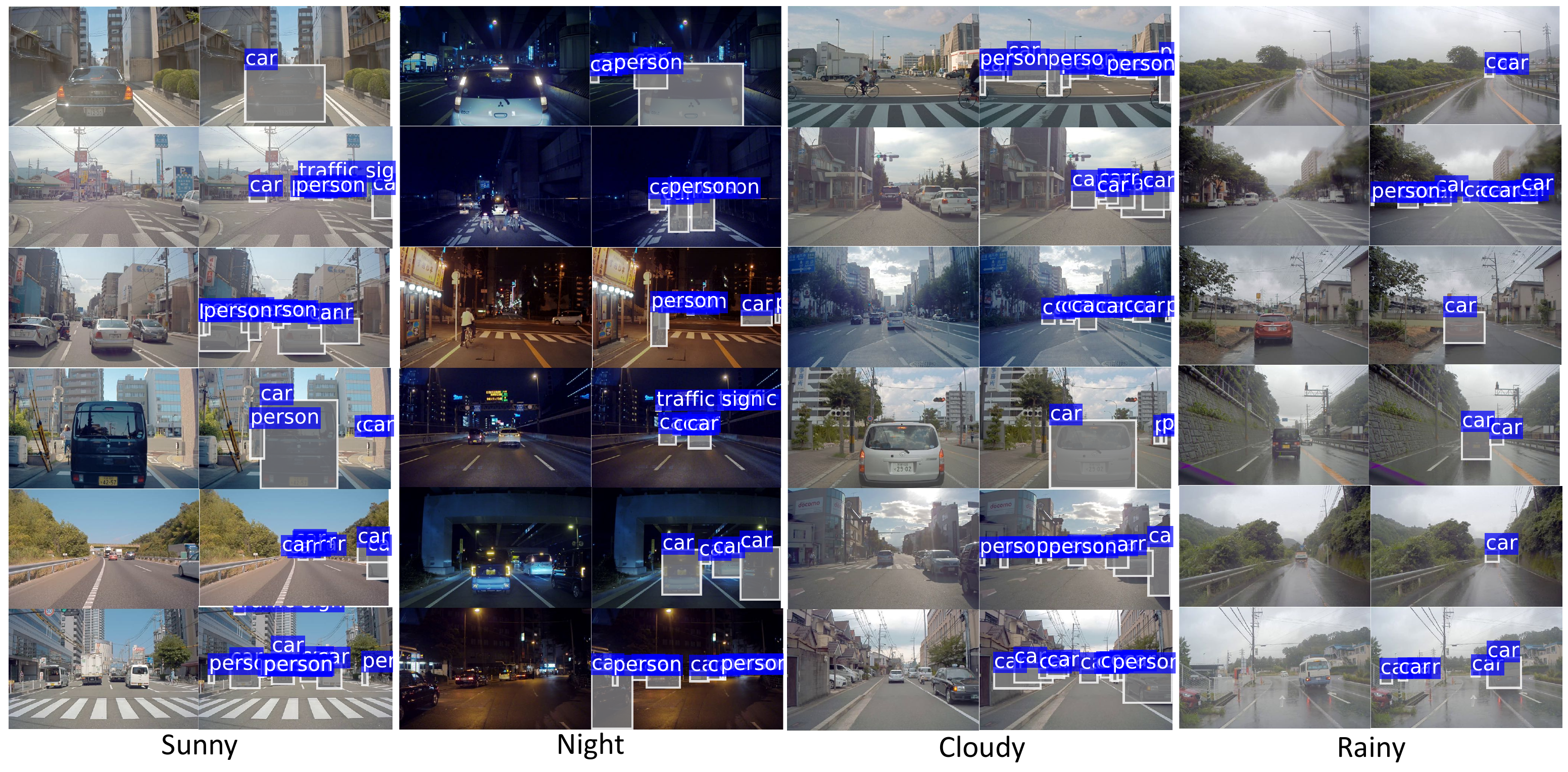}
	\vspace{-0.08in}
	\caption{Image samples from our benchmark grouped by their domain categories (sunny, night, cloudy and rainy). In each group, left are original images and right are images with corresponding bounding box annotations.} 
	\label{benchmark}
	\vspace{-0.1in}
\end{figure*}

\noindent{\textbf{A New Benchmark for Unpaired Image-to-Image Translation.}}
 We introduce a new large-scale street scene centric dataset that addresses three core research problems in I2I translation: (1) unsupervised learning paradigm, meaning that there is no specific one-to-one mapping in the dataset; (2) multimodal domains incorporation. Most existing I2I translation datasets provide only two different domains, which limit the potential to explore more challenging task like multi-domain incorporation  circumstance. Our dataset contains four domains: sunny, night, cloudy and rainy\footnote{For safety, we collect the rainy images after the rain, so this category looks more like overcast weather with wet road (see Fig.~\ref{benchmark} for more details).} in a unified street scene; and (3) multi-granularity (global and instance-level) information. Our dataset provides instance-level bounding box annotations, which can utilize more details for learning a translation model.
Tab.~\ref{feature} shows a feature-by-feature comparison among various I2I translation datasets. We also visualize some examples of the dataset in Fig.~\ref{benchmark}. For instance category, we annotate three common objects in street scenes including: car, person,  traffic sign (speed limited sign). The detailed statistics (\# images) of the entire dataset are shown in Sec.~\ref{data}.

\section{Instance-aware Image-to-Image Translation} \label{method}
\begin{figure*}[t]
	\centering
	\includegraphics[width=0.85\textwidth]{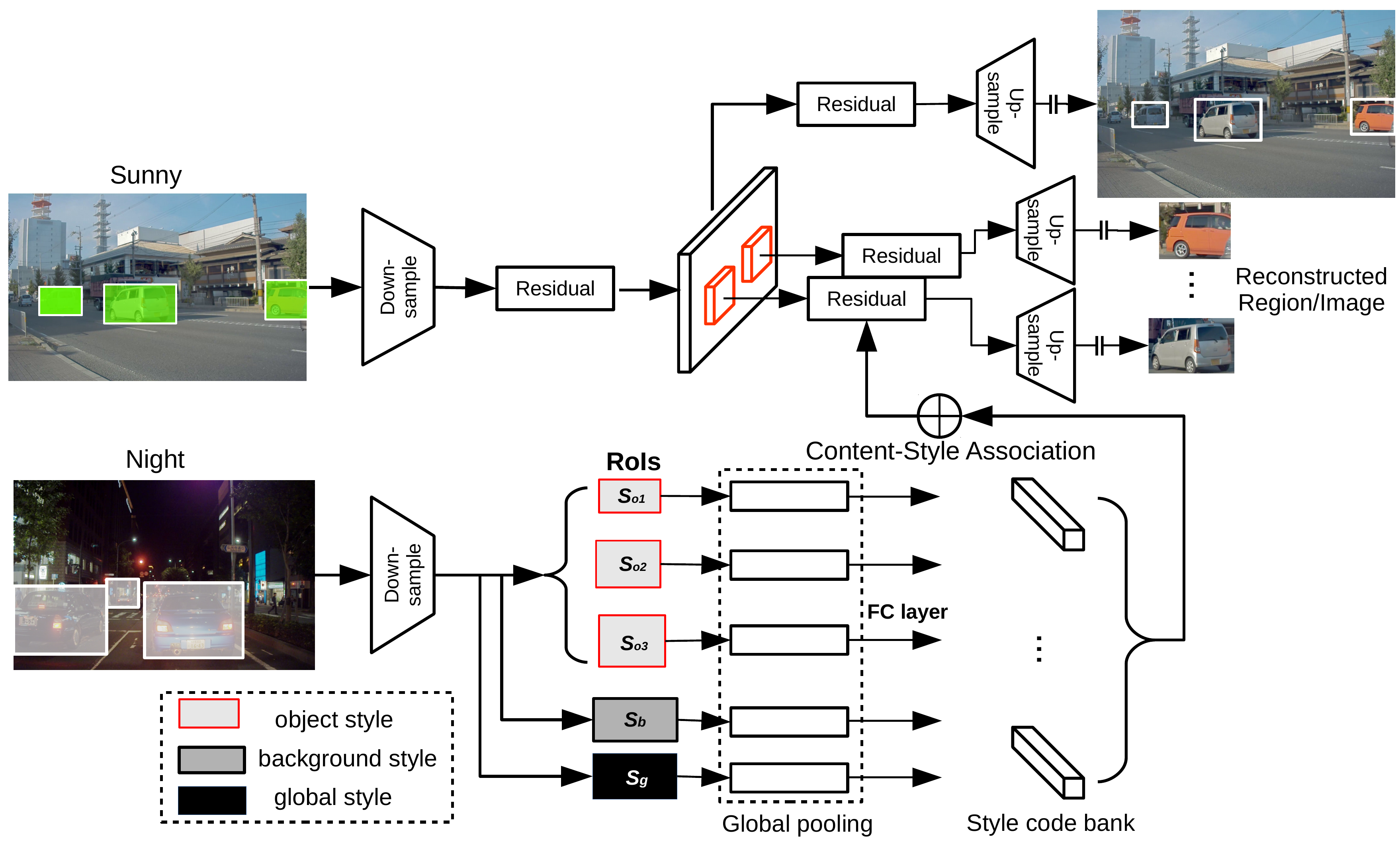}
	\caption{Overview of our instance-aware cross-domain I2I translation. The whole framework is based on the MUNIT method~\cite{huang2018multimodal}, while we further extend it to realize the instance-level translation purpose. Note that after content-style association, the generated images will place in the target domain, so a translation back process will be employed before self-reconstruction, which is not illustrated here.} 
	\label{overview}
	\vspace{-0.07in}
\end{figure*}

\begin{figure}[h]
	\centering
	\includegraphics[width=0.45\textwidth]{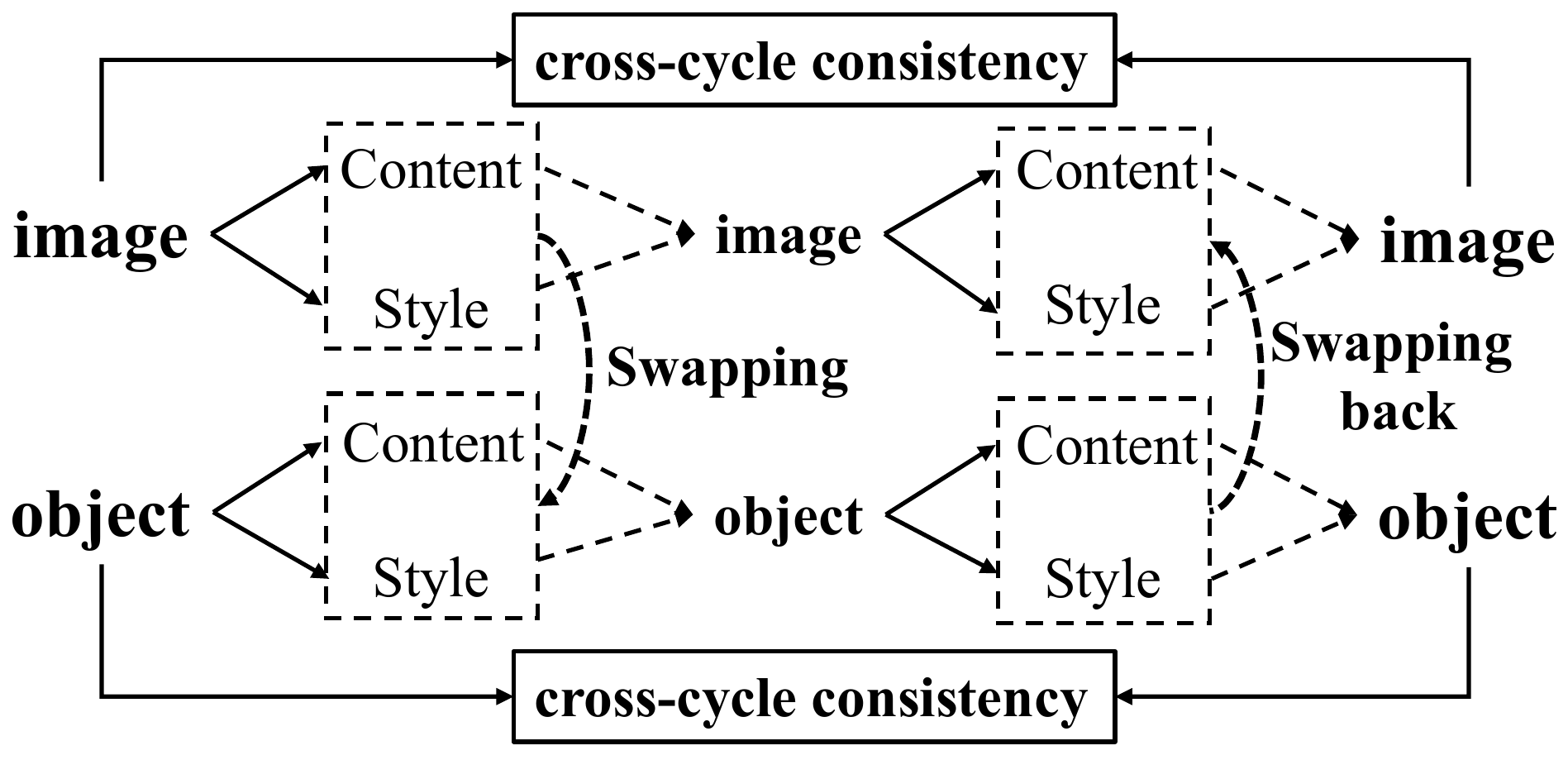}
	\caption{Illustration of our cross-cycle consistency process. We only show cross-granularity (image $\leftrightarrow$ object), the cross-domain consistency ($\mathcal X \leftrightarrow \mathcal Y$) is similar to the above paradigm.} 
	\label{cross_cycle}
	\vspace{-0.1in}
\end{figure}

Our goal is to realize the instance-aware I2I translation between two different domains without paired training examples. 
We build our framework by leveraging the MUNIT~\cite{huang2018multimodal} and DRIT~\cite{lee2018diverse} methods. To avoid repetition, we omit some innocuous details. Similar to MUNIT~\cite{huang2018multimodal} and DRIT~\cite{lee2018diverse}, our method is straight-forward and simple to implement. As illustrated in Fig.~\ref{cross_cycle}, our translation model consists of two encoders $E_g, E_o$ ($g$ and $o$ denote the global and instance image regions respectively), and two decoders $G_g,G_o$ in each domain $\mathcal X$ or $\mathcal Y$. A more detailed illustration is shown in Fig.~\ref{overview}. Since we have the object coordinates, we can crop the object areas and feed them into the instance-level encoder to extra the content/style vectors. An alternative method for object content vectors is to adopt RoI pooling~\cite{girshick2015fast} from the global image content features. Here we use image crop (object region) and share the parameters for the two encoders, which is more easier to implement.

\begin{figure}[t]
	\centering
	\includegraphics[width=0.48\textwidth]{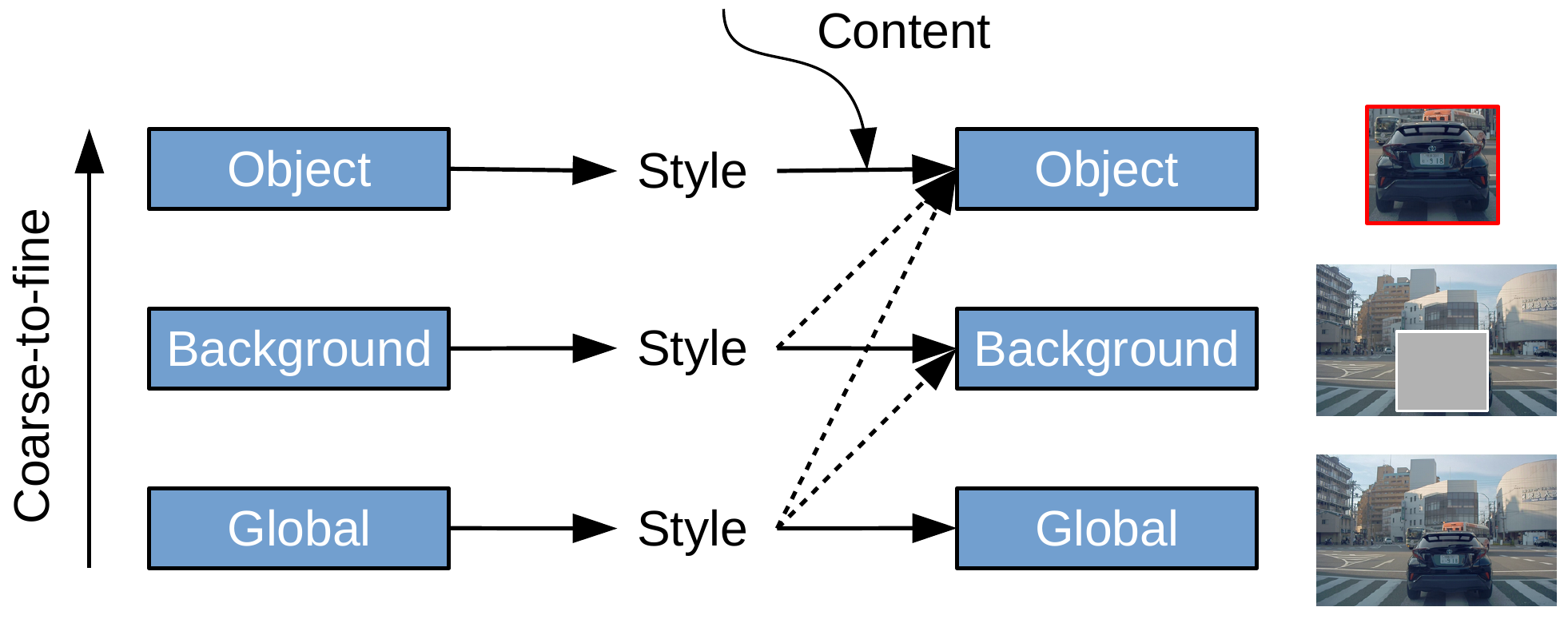}
	\caption{Our content-style pair association strategy. Only coarse styles can be applied to fine contents, the reversal of processing flow is not allowed during training.} 
	\label{strategy}
	\vspace{-0.16in}
\end{figure}


\noindent{\textbf{Disentangle content and style on object and entire image.}}
As ~\cite{cheung2014discovering,mathieu2016disentangling,huang2018multimodal,lee2018diverse}, our method also decomposes input images/objects into a shared content space and a domain-specific style space.
 Take global image as an example, each encode $E_g$ can decompose the input to a content code $c_{g}$ and a style code $s_{g}$, where 
$E_{g}=(E_{g}^c,E_{g}^s)$, \ $c_{g}=E_{g}^c(I)$, \ $s_{g}=E_{g}^s(I)$, $I$ denotes the input image representation. $c_g$ and $s_g$ are global-level content/style features.

\noindent{\textbf{Generate style code bank.}}
We generate the style codes from objects, background and entire images, which form our style code bank for the following swapping operation and translation. In contrast, MUNIT~\cite{huang2018multimodal} and DRIT~\cite{lee2018diverse} use only the entire image style or attribute, which is struggling to model and cover the rich image spatial representation.

\noindent{\textbf{Associate content-style pairs for cyclic reconstruction.}} 
Our cross-cycle consistency is performed by swapping encoder-decoder pairs (dashed arc lines in Fig.~\ref{cross_cycle}). The cross-cycle includes two modes: cross-domain ($\mathcal X \leftrightarrow \mathcal Y$) and cross-granularity (entire image $\leftrightarrow$ object). 
We illustrate cross-granularity (image $\leftrightarrow$ object) in Fig.~\ref{cross_cycle}, the cross-domain consistency ($\mathcal X \leftrightarrow \mathcal Y$) is similar to MUNIT~\cite{huang2018multimodal} and DRIT~\cite{lee2018diverse}.
As shown in Fig.~\ref{strategy}, the swapping or content-style association strategy is a hierarchical structure across multi-granularity areas. Intuitively, the coarse (global) style can affect fine content and be adopted to local areas, while it's not true if the process is reversed. Following~\cite{huang2018multimodal}, we also use AdaIN~\cite{huang2017arbitrary} to combine the content and style vectors.

\noindent{\textbf{Incorporate Multi-Scale.}}
It's technically easy to incorporate multi-scale advantage into the framework. We simply replace the object branch in Fig.~\ref{cross_cycle} with resolution-reduced images.  In our experiments, we use 1/2 scale and original size images as pairs to perform scale-augmented training. Specifically, styles from small size and original size images can be performed to each other, and the generator needs to learn multi-scale reconstruction for both of them, which leads to more accurate results.

\noindent{\textbf{Reconstruction loss.}}
We use self-reconstruction and cross-cycle consistency loss~\cite{lee2018diverse} for both entire image and object that encourage reconstruction of them. With encoded $c$ and $s$, the decoders should decode them back to original input,
\begin{equation}
\hat I = {G_g}(E_g^c(I),\;E_g^s(I)),\;\hat o = {G_o}(E_o^c(o),\;E_o^s(o))
\end{equation}
We can also reconstruct the latent distribution (i.e. content and style vectors) as~\cite{huang2018multimodal}.
\begin{equation}
{\hat c}_o = E_o^c({G_o}({c_o},{s_g})),\;{{\hat s}_o} = E_o^s({G_o}({c_o},{s_g}))
\end{equation}
where $c_o$ and $s_g$ are instance-level content and global-level style features.
Then, we can use the following formation to learn a reconstruction of them:
\begin{equation}\label{recon}
\mathcal{L}_{recon}^k = {\mathbb{E}_{k \sim p(k)}}\left[ {{{\left\| {\hat k - k} \right\|}_1}} \right]
\end{equation}
where $k$ can be $I$, $o$, $c$ or $s$. $p(k)$ denotes the distribution of data k.
The formation of cross-cycle consistency is similar to this process and more details can be referred to ~\cite{lee2018diverse}.

\noindent{\textbf{Adversarial loss.}}
Generative adversarial learning~\cite{goodfellow2014generative} has been adapted to many visual tasks, e.g., detection~\cite{nguyen2017shadow,bai2018finding}, inpainting
~\cite{pathak2016context,Yu_2018_CVPR,iizuka2017globally,yu2018free}, ensembling~\cite{shen2019MEAL}, etc. We adopt adversarial loss $\mathcal{L}_{adv}$ where $D^g_\mathcal{X}$, $D^o_\mathcal{X}$, $D^g_\mathcal{Y}$ and $D^o_\mathcal{Y}$ attempt to discriminate between real and synthetic images/objects in each domain.
We explore two designs for the discriminators: weight-sharing or weight-independent for global and instance images in each domain. The ablation experimental results are shown in Tab.~\ref{LPIPS} and Tab.~\ref{IS}, we observe that shared discriminator is a better choice in our experiments.

\noindent{\textbf{Full objective function.}}
The full objective function of our framework is:
\begin{equation}
\begin{gathered}
  \mathop {\min }\limits_{{E_\mathcal{X}},{E_\mathcal{Y}},{G_\mathcal{X}},{G_\mathcal{Y}}} \mathop {\max }\limits_{{D_\mathcal{X}},{D_\mathcal{Y}}} \mathcal{L}({E_\mathcal{X}},{E_\mathcal{Y}},{G_\mathcal{X}},{G_\mathcal{Y}},{D_\mathcal{X}},{D_\mathcal{Y}}) \hfill \\
   = \underbrace {{\lambda _g}(\mathcal{L}_{}^{{g_\mathcal{X}}} + \mathcal{L}_{}^{{g_\mathcal{Y}}}) + {\lambda _{{c_g}}}(\mathcal{L}_g^{{c_\mathcal{X}}} + \mathcal{L}_g^{{c_\mathcal{Y}}}) + {\lambda _{{s_g}}}(\mathcal{L}_g^{{s_\mathcal{X}}} + \mathcal{L}_g^{{s_\mathcal{Y}}})}_{global - level\;reconstruction\;loss} \hfill \\
  \quad  + \underbrace {{\lambda _o}(\mathcal{L}_{}^{{o_\mathcal{X}}} + \mathcal{L}_{}^{{o_\mathcal{Y}}}) + {\lambda _{{c_o}}}(\mathcal{L}_o^{{c_\mathcal{X}}} + \mathcal{L}_o^{{c_\mathcal{Y}}}) + {\lambda _{{s_o}}}(\mathcal{L}_o^{{s_\mathcal{X}}} + \mathcal{L}_o^{{s_\mathcal{Y}}})}_{instance - level\;reconstruction\;loss} \hfill \\
  \quad  + \underbrace {\mathcal{L}_{adv}^{{\mathcal{X}_g}} + \mathcal{L}_{adv}^{{\mathcal{Y}_g}}}_{global - level\;GAN\;loss} + \underbrace {\mathcal{L}_{adv}^{{\mathcal{X}_o}} + \mathcal{L}_{adv}^{{\mathcal{Y}_o}}}_{instance - level\;GAN\;loss} \hfill \\ 
\end{gathered} 
\end{equation}

During inference time, we simply use the global branch to generate the target domain images (See Fig.~\ref{overview} upper-right part) so that it's not necessary to use bounding box annotations at this stage, and this strategy can also guarantee that the generated images are harmonious for objects and background.

\begin{table}[t]
	\centering
	\resizebox{0.5\textwidth}{!}{%
		\begin{tabular}{l|c|c|c}
			\hline
			\bf Domain & \bf Training (85\%) &\bf Testing (15\%) & \bf Total (100\%)  \\ \hline\hline
			\bf Sunny  & 49,663           & 8,764           & 58,427  \\ \hline
			\bf Night  & 24,559           & 4,333           & 28,892  \\ \hline
			\bf Rainy  & 6,041            & 1,066           & 7,107   \\ \hline
			\bf Cloudy & 51,938           & 9,165           & 61,103  \\ \hline 
			\bf Total  & 132,201          & 23,328          & 155,529 \\ \hline
		\end{tabular}
	}
	\caption{Statistics (\# images) of the entire dataset across four domains: sunny, night, rainy and cloudy. The data is divided into two subsets: 85\% for training and 15\% for testing.}
	\label{statistics}
	\vspace{-2.0ex}
\end{table}

\begin{table*}[t]
	\centering
	\begin{tabular}{l|c|c|c|c}
		\hline
		\bf Method      &\multicolumn{4}{c}{\bf Diversity} \\ \hline 
		&  sunny $\to$ night &sunny$\to$rainy & sunny$\to$cloudy & Average\\ \hline \hline
		UNIT~\cite{liu2017unsupervised}        &  0.067   &   0.062&  0.068 &  0.066    \\ \hline
		CycleGAN~\cite{CycleGAN2017}        &   0.016  &  0.008 &  0.011 &   0.012   \\ \hline
		MUNIT~\cite{huang2018multimodal}       &  0.292  & 0.239  &  0.211  &    0.247   \\ \hline
		DRIT~\cite{lee2018diverse}        & 0.231    &0.173&0.166&  0.190    \\ \hline \hline
		INIT  w/ D$_s$      &  \bf 0.330   & \bf  0.267   &  \bf  0.224  & \bf 0.274   \\ \hline
		INIT  w/o D$_s$     & 0.324     &    0.238  &  0.177   &   0.246  \\ \hline\hline
		Real Images &  0.573   &  0.489&  0.465  &  0.509    \\ \hline
	\end{tabular}
	\caption{{\bf Diversity scores on our dataset.} We use the average LPIPS distance~\cite{zhang2018unreasonable} to measure the diversity of generated images.}
	\label{LPIPS}
	\vspace{-1.0ex}
\end{table*}

{	\renewcommand{\arraystretch}{1.13}
	\setlength{\tabcolsep}{1.2em}
	\begin{table*}[t]
		\centering
		\resizebox{1\textwidth}{!}{%
			\begin{tabular}{l|c|c|c|c|c|c|c|c|c|c|c|c}
				\hline
				&\multicolumn{2}{c|}{CycleGAN~\cite{CycleGAN2017} }  & \multicolumn{2}{c|}{UNIT~\cite{liu2017unsupervised} } & \multicolumn{2}{c|}{MUNIT~\cite{huang2018multimodal} } & \multicolumn{2}{c|}{DRIT~\cite{lee2018diverse}} & \multicolumn{2}{c|}{INIT w/ D$_s$} & \multicolumn{2}{c}{INIT w/o D$_s$}  \\ \hline
				& CIS          & IS & CIS          & IS         & CIS          & IS          & CIS          & IS         & CIS          & IS    & CIS & IS    \\ \hline
				sunny$ \to $night &0.014   & 1.026  &  0.082  &   1.030  &    1.159   &    1.278    &    1.058     & 1.224    &   1.060  &   1.118    &1.083 &   1.120  \\ 
				night$ \to $sunny & 0.012 & 1.023  &  0.027  &   1.024 &    1.036     &     1.051     &  1.024     &  1.099   &   1.045   &    1.080    & 1.024 &  1.104    \\ \hline
				sunny$ \to $rainy & 0.011 & 1.073 & 0.097   &   1.075   &   1.012    &   1.146   &  1.007   &  1.207  &  1.036  &   1.152   &1.034 &   1.146   \\ 
				rainy$ \to $sunny & 0.010  & 1.090  &  0.014 &  1.023   &     1.055   &   1.102    &  1.028   &   1.103   &    1.060     &   1.119   &1.059 &  1.124    \\ \hline
				sunny$ \to $cloudy  & 0.014 &  1.097 & 0.081  & 1.134    &    1.008    &   1.095   &   1.025    &   1.104  &   1.040    &  1.142   & 1.025&  1.147     \\ 
				cloudy$ \to $sunny  &  0.090  &  1.033 & 0.219  &   1.046   &    1.026    &  1.321   &    1.046  &  1.249    &    1.016     & 1.460    & 1.006 &  1.363     \\ \hline
				Average   & 0.025 &  1.057 & 0.087 &  1.055   &   1.032   & 1.166   &   1.031  & 1.164  & \bf  1.043   & \bf 1.179    & 1.039&    1.167  \\ \hline
			\end{tabular}
		}
		\caption{{\bf Comparison of Conditional Inception Score (CIS) and Inception Score (IS).} To obtain high CIS and IS scores, a model is required to synthesis images that are more realistic, diverse with high-quality.}
		\label{IS}
	\end{table*}
}

\begin{table*}[]
	\centering
	\begin{tabular}{cc|cc|ccc|ccc}
		\hline
		\multicolumn{2}{c|}{COCO 2017 training }   & \multicolumn{2}{c|}{COCO 2017 validation}              &\multicolumn{3}{c|}{object detection (\%)}    &\multicolumn{3}{c}{instance segmentation (\%)}     \\ \hline 
		\multirow{2}{*}{Real}  &   \multirow{2}{*}{Synthetic}        & \multirow{2}{*}{Real}     &     \multirow{2}{*}{Synthetic}   & \multicolumn{3}{c|}{Avg. Precision, IoU:}    & \multicolumn{3}{c}{Avg. Precision, mask:}  \\
		&            &     &      & 0.5:0.95        & 0.5        & 0.75     & 0.5:0.95        & 0.5        & 0.75            \\ \hline
		\Checkmark   &                         & \Checkmark    &                       &   37.7     &   59.2    &  40.8  &    34.3    &  56.0     &    36.2     \\
		\Checkmark &                           &                       & \Checkmark    &  30.4     & 49.7 & 32.6 & 27.8 & 46.6 & 29.2  \\
		&    \Checkmark     &  \Checkmark   &                       &  30.0   &  50.0    &   31.6   &    27.2    &   46.5    & 28.0     \\
		& \Checkmark       &                       &    \Checkmark  &   30.5     &   49.7    &  32.7    &    27.8    &   46.4    &   29.0   \\ \hline
		\Checkmark  &  \Checkmark    &                        &  \Checkmark  &     32.6$^{\uparrow  2.1}$     &   52.6$^{\uparrow  2.9}$   &  34.2$^{\uparrow  1.5}$ &  29.0$^{\uparrow  1.2}$   & 49.0$^{\uparrow  2.6}$  & 29.8$^{\uparrow  0.8}$  \\ 
		\Checkmark  &  \Checkmark    &  \Checkmark   &                       &     38.8$^{\uparrow  1.1}$   &    60.2$^{\uparrow  1.0}$     &   42.5$^{\uparrow  1.7}$  &    35.2$^{\uparrow  0.9}$    &   57.0$^{\uparrow  1.0}$    &   37.4$^{\uparrow 1.2}$   \\
		\hline
	\end{tabular}
	\caption{\textbf{Mask-RCNN with ResNet-50-FPN~\cite{lin2017feature} detection and segmentation results on MS COCO 2017 \texttt{val} set.}}
	\label{COCO}
		\vspace{-0.1in}
\end{table*}

\section{Experiments and Analysis }\label{data}

We conduct experiments on our collected dataset (INIT). We also use COCO dataset~\cite{lin2014microsoft} to verify the effectiveness of data augmentation.

\noindent{\textbf{INIT Dataset.}}
INIT dataset consists of 132,201 images for training and 23,328 images for testing. The detailed statistics are shown in Tab.~\ref{statistics}. All the data are collected in Tokyo, Japan with SEKONIX AR0231 camera.  The whole collection process lasted about three months.

\noindent{\textbf{Implementation Details.}}
Our implementation is based on MUNIT\footnote{https://github.com/NVlabs/MUNIT} with PyTorch~\cite{paszke2017automatic}. For I2I translation, we resize the short side of images to 360 pixels due to the limitation of GPU memory. For COCO image synthesis, since the training images (INIT dataset) and target images (COCO) are in different distributions, we keep the original size of our training image and crop 360$\times$360 pixels to train our model, in order to learn more details of images and objects, meanwhile, ignore the global information. In this circumstance, we build our object part as an independent branch and each object is resized to 120$\times$120 pixels during training.

\subsection{Baselines}
We perform our evaluation on the following four recent proposed state-of-the-art unpaired I2I translation methods:
\begin{itemize}[leftmargin=0.3cm]
	\addtolength{\itemsep}{-0.05in}
	\item[-] CycleGAN~\cite{CycleGAN2017}: CycleGAN contains two translation functions ($\mathcal X \rightarrow \mathcal Y$ and $\mathcal X \leftarrow \mathcal Y$), and the corresponding adversarial loss. It assumes that the input images can be translated to another domain and then can be mapped back with a cycle consistency loss.
	\item[-] UNIT~\cite{liu2017unsupervised}: The UNIT method is  an extension of CycleGAN~\cite{CycleGAN2017} that is based on the shared latent space assumption. It contains two VAE-GANs and also uses cycle-consistency loss~\cite{CycleGAN2017} for learning models.
	\item[-] MUNIT~\cite{huang2018multimodal}: MUNIT consists of an encoder and a decoder for each domain. It assumes that the image representation can be decomposed into a domain-invariant content space and a domain-specific style space. The latent vectors of each encoder are disentangled to a content vector and a style vector. I2I translation is performed by swapping content-style pairs.
	\item[-] DRIT~\cite{lee2018diverse}: The motivation of DRIT is similar to MUNIT. It consists of content encoders, attribute encoders, generators and domain discriminators for both domains. The content encoder maps images into a shared content space and the attribute encoder maps images into a domain-specific attribute space. A cross-cycle consistency loss is adopted for performing I2I translation.
\end{itemize}

\subsection{Evaluation}
We adopt the same evaluation protocol from previous unsupervised I2I translation works and evaluate our method with the LPIPS Metric~\cite{zhang2018unreasonable}, Inception Score (IS)~\cite{salimans2016improved} and Conditional Inception Score (CIS)~\cite{huang2018multimodal}.

\begin{figure}[t]
	\centering
	\includegraphics[width=0.48\textwidth]{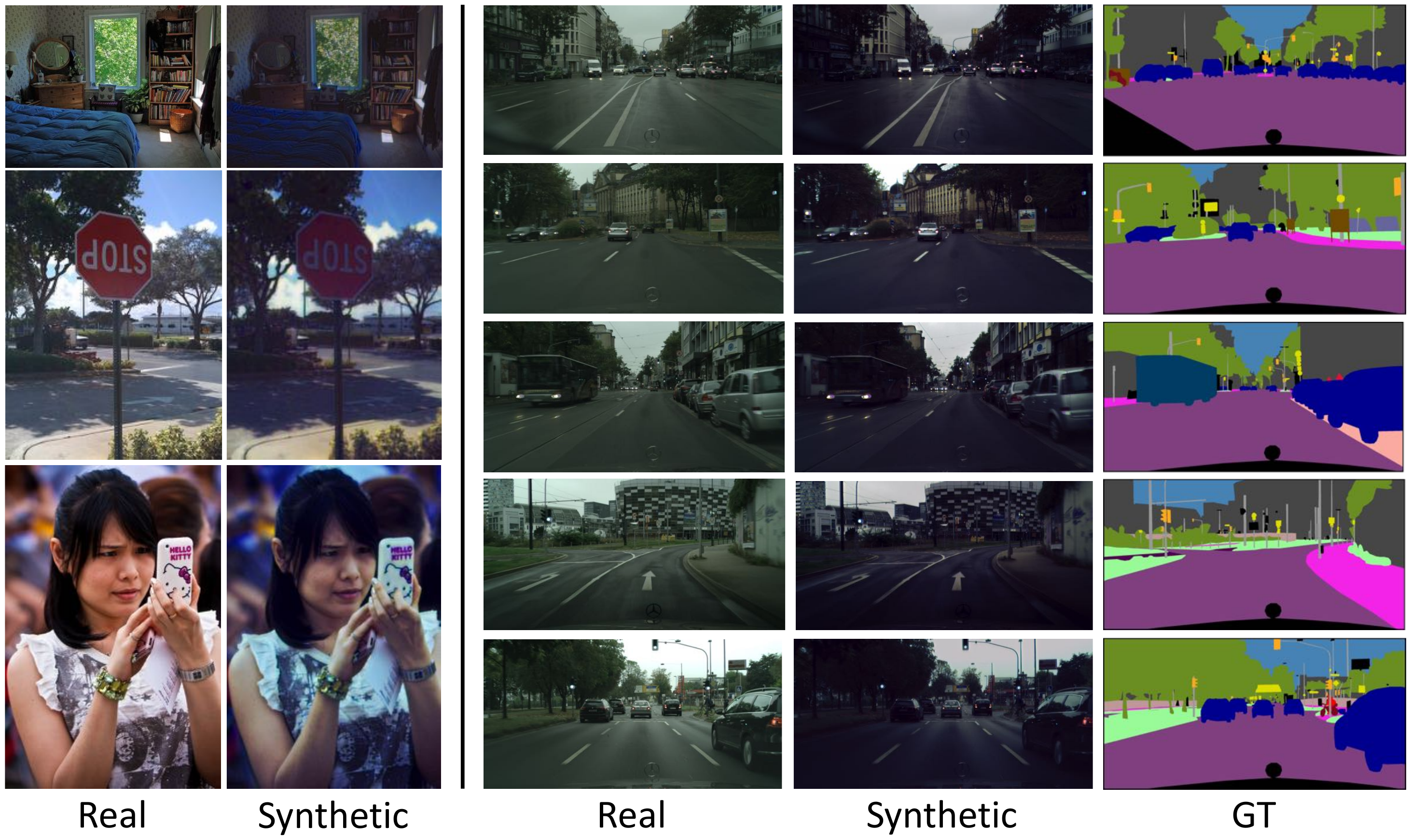}
	\caption{{\bf Visualization of our synthetic images.} The left group images are from COCO and the right are from Cityscapes.} 
	\label{synth}
	\vspace{-0.16in}
\end{figure}

\noindent{\textbf{LPIPS Metric.}}
Zhang et al. proposed LPIPS distance~\cite{zhang2018unreasonable} to measure the translation diversity, which has been verified to correlate well with human perceptual psychophysical similarity. Following~\cite{huang2018multimodal}, we calculate the average LPIPS distance between 19 pairs of randomly sampled translation outputs from 100 input images of our test set. Following~\cite{huang2018multimodal} and recommended by~\cite{zhang2018unreasonable}, we also use the pre-trained AlexNet~\cite{krizhevsky2012imagenet} to extract deep features.

Results are summarized in Tab.~\ref{LPIPS}, ``INIT w/ D$_s$'' denotes we train our model with shared discriminator between entire image and object. ``INIT w/o D$_s$'' denotes we build separate discriminators for image and object. Thanks to the coarse and fine styles we used, our average INIT w/ D$_s$ score outperforms MUNIT with a  notable margin. We also observe that our dataset (real image) has a very large diversity score, which indicates that the dataset is diverse and challenging.

\begin{figure}[t]
	\centering
	\includegraphics[width=0.45\textwidth]{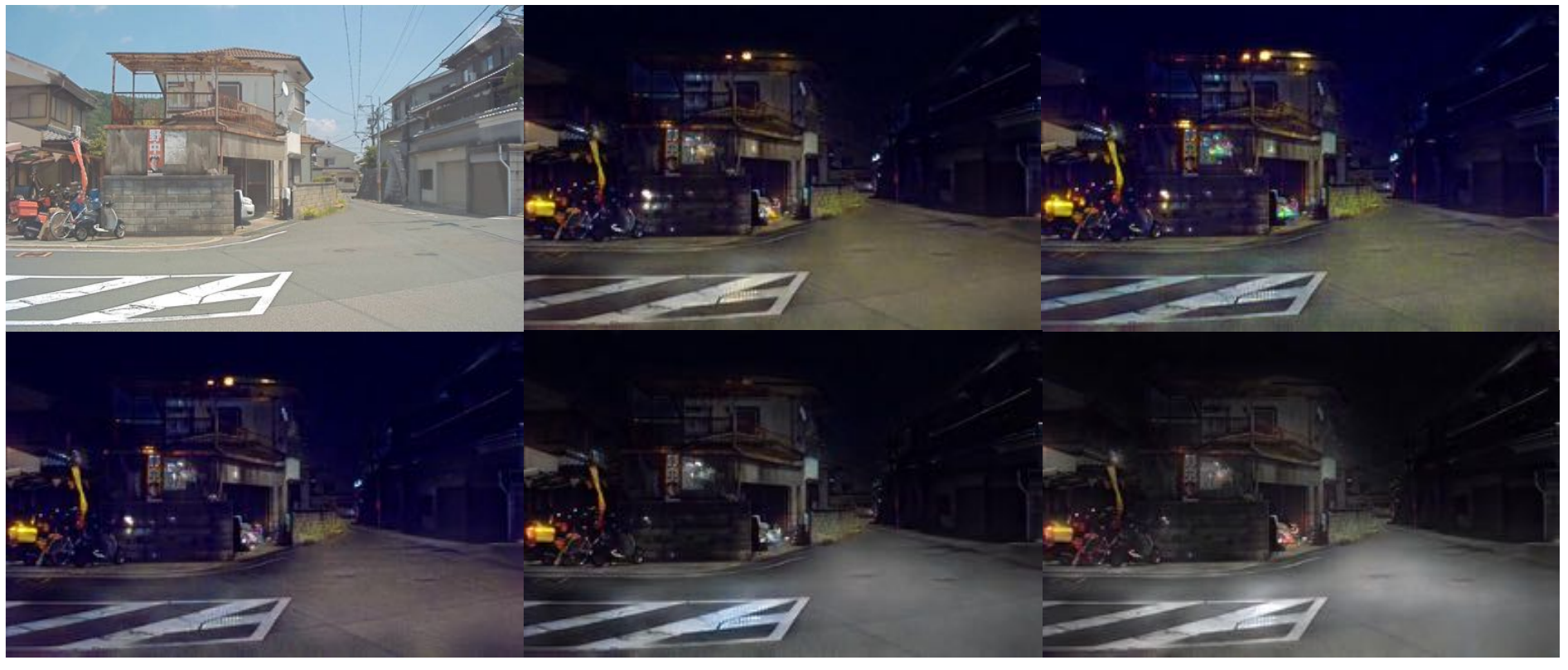}
	\caption{{\bf Visualization of multimodal results.} We use randomly sampled style codes to generate these images and the darkness are slightly different across them.} 
	\label{multimodal}
	\vspace{-0.1in}
\end{figure}

\begin{table}[t]
	\centering
	\begin{tabular}{l|c|c|c}
		\hline
		COCO 2017 (\%)          & IoU & IoU$_{0.5}$ & IoU$_{0.75}$ \\ \hline \hline
		+Syn. (MUNIT~\cite{huang2018multimodal}) &   +0.7  &   +0.4     &   +1.0      \\ 
		+Syn. (Ours)                                                 &   \bf +1.1  & \bf   +1.0    & \bf  + 1.7     \\ \hline
	\end{tabular}
	\caption{Improvement comparison on COCO detection with different image synthetic methods.}
	\label{improvement}
	\vspace{-0.1in}
\end{table}

\begin{table}[t]
	\centering
	\begin{tabular}{l|c|c}
		\hline
		&    Metric   &  Percentage (\%) \\ \hline \hline
		COCO &      Det.\&Seg.      &  $ \downarrow $19.1 \& $ \downarrow $19.0        \\ 
		Cityscapes    &   mIoU\&mAcc   & \bf $ \downarrow $ 2.6 \& $ \downarrow $2.4    \\ \hline
	\end{tabular}
	\vspace{0.1in}
	\caption{{\bf Performance decline} when training and testing on real image, and comparing to results on synthetic image. We adopt PSPNet~\cite{zhao2017pyramid} with ResNet-50~\cite{he2016deep} on Cityscapes~\cite{cordts2016cityscapes} and obtain (real\&real): mIoU: 76.6\%, mAcc: 83.1\%;  (syn.\&syn.): 74.6\%/81.1\% .}
	\label{decline}
	\vspace{-0.21in}
\end{table}

\noindent{\textbf{Inception Score (IS) and Conditional Inception Score (CIS).}}
We use the Inception Score (IS)~\cite{salimans2016improved} and Conditional Inception Score (CIS)~\cite{huang2018multimodal} to evaluate our learned models. IS measures the diversity of all output images and CIS measures diversity of output conditioned on a single input image, which is a modified IS that is more suitable for evaluating multimodal I2I translation task. The detailed definition of CIS can be referred to~\cite{huang2018multimodal}. We also employ with Inception V3 model~\cite{szegedy2016rethinking} to fine-tune our classification model on four domain category labels of our dataset. Other settings are the same as ~\cite{huang2018multimodal}. It can be seen in Tab.~\ref{IS} that our results are consistently better than the baselines MUNIT and DRIT.

\begin{figure}[t]
	\centering
	\includegraphics[width=0.48\textwidth]{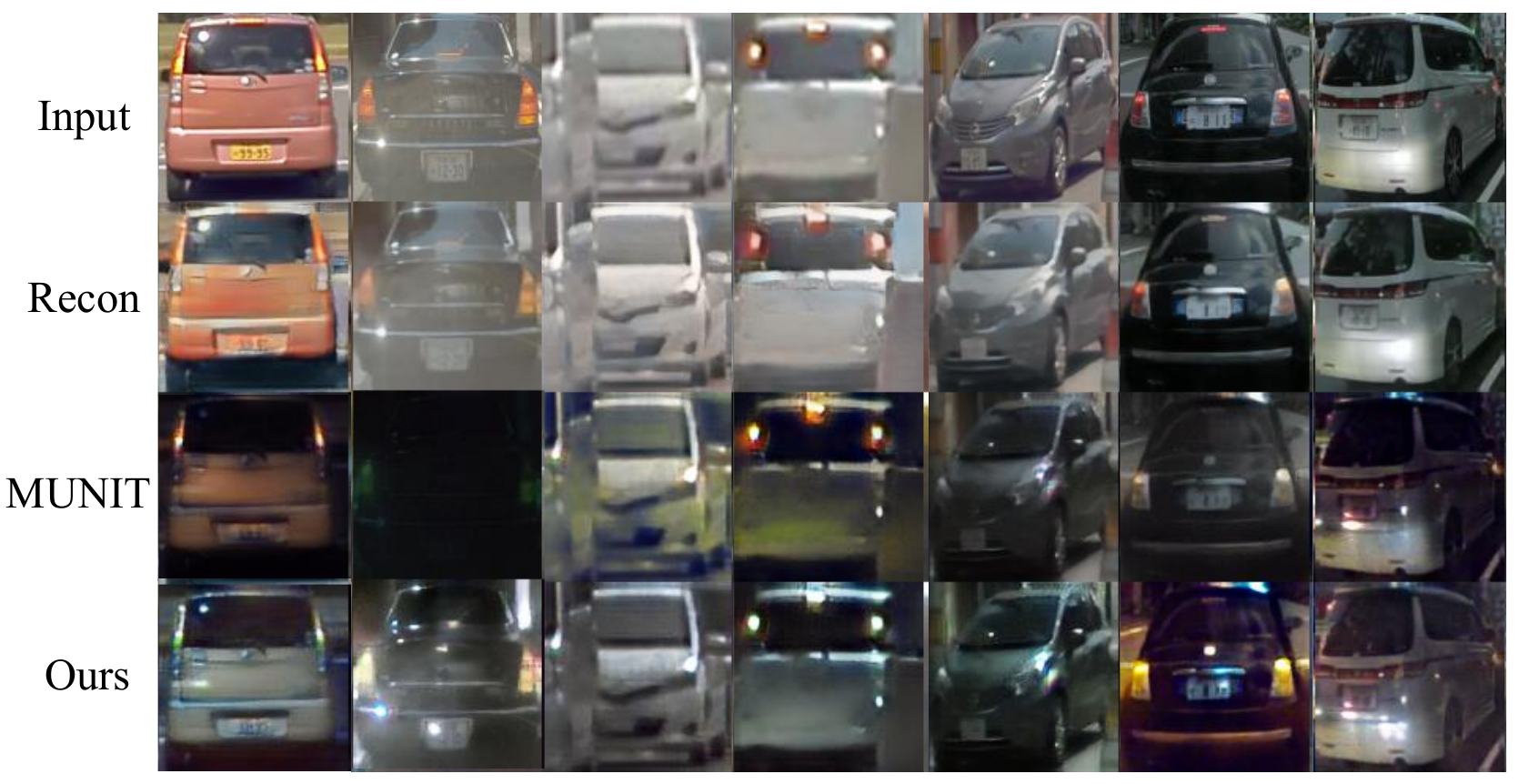}
	\caption{{\bf Qualitative comparison on randomly selected instance level results.} The first row shows the input objects. The second row shows the self-reconstruction results. The third and fourth rows show outputs from MUNIT and ours, respectively.} 
	\label{instance}
	\vspace{-0.16in}
\end{figure}

\begin{figure*}[t]
	\centering
	\includegraphics[width=0.9\textwidth]{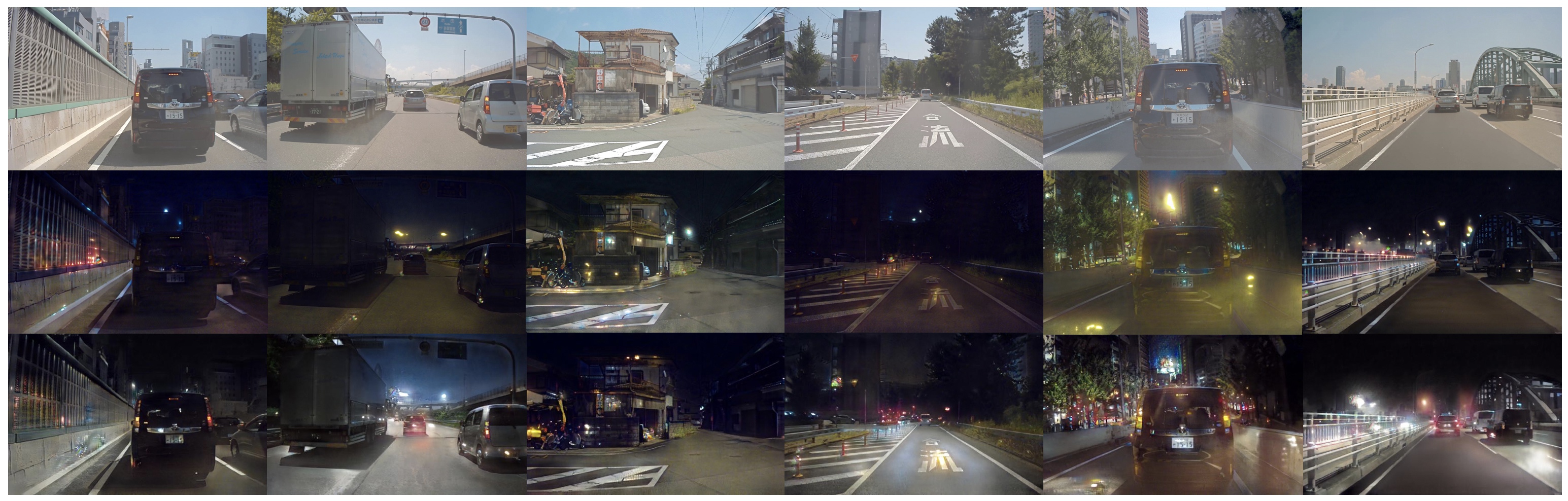}
		\vspace{-0.03in}
	\caption{{\bf Case-by-case  comparison on sunny$\to$night.} The first row shows the input images. The second and third rows show random outputs from MUNIT~\cite{huang2018multimodal} and ours, respectively.} 
	\label{dark}
	\vspace{-0.17in}
\end{figure*}

\noindent{\textbf{Image Synthesis on Multiple Datasets}}
The visualization of our synthetic images is shown in Fig.~\ref{synth}. The left group images are on COCO and the right are on Cityscapes. We observe that the most challenging problem for multiple datasets synthesis is the inter-class variance among them.

\noindent{\textbf{Data Augmentation for Detection \& Segmentation on COCO.}}
We use Mask RCNN~\cite{he2017mask} framework for the experiments. A synthetic copy of entire COCO dataset is generated by our sunny$\to$night model. We employ open-source implementation of Mask RCNN\footnote{https://github.com/facebookresearch/maskrcnn-benchmark} for training the COCO models. For training, we use the same number of training epochs and other default settings including the learning rating schedule, \# batchsize, etc.

All results are summarized in Tab.~\ref{COCO}, the first column (group) shows the training data we used, the second group shows the validation data where we tested on. The third and fourth groups are detection and segmentation results, respectively. We can observe that our real-image trained model can obtain 30.4\% mAP on synthetic validation images, this indicates that the distribution differences between original COCO and our synthetic images are not very huge. It seems that our generation process is more likely to do photo-metric distortions or brightness adjustment of images, which can be regarded as a data augmentation technique and has been verified the effectiveness for object detection in~\cite{liu2016ssd}. From the last two rows we can see that not only the synthetic images can help improve the real image testing performance, but the real image can also boost the results of synthetic images (both train and test on synthetic images). 
We also compare improvement with different generation methods in Tab.~\ref{improvement}. The results show that our object branch can bring more benefits for detection task than the baseline. We also believe that the proposed data augmentation method can benefit to some limited training data scenarios like learning detectors from scratch~\cite{shen2017dsod,law2018cornernet,he2018rethinking,duan2019CenterNet}.

We further conduct scene parsing on Cityscapes~\cite{cordts2016cityscapes}. However, we didn't see obvious improvement in this experiment. Using PSPNet~\cite{zhao2017pyramid} with ResNet-50~\cite{he2016deep}, we obtain mIoU: 76.6\%, mAcc: 83.1\% when training and testing on real images and 74.6\%/81.1\% on both synthetic images. We can see that the gaps between real and synthetic image are really small. We conjecture this case (no gain) is because 
the synthetic Cityscapes is too close to the original one. We compare the performance decline in Tab.~\ref{decline}. Since the metrics are different in COCO and Cityscapes, we use the relative percentage for comparison. The results indicate that the synthetic images may be more diverse for COCO since the decline is much smaller on Cityscapes.

\section{Analysis}

\noindent{\textbf{Qualitative Comparison.}}
 We qualitatively compare our method with baseline MUNIT~\cite{huang2018multimodal}. Fig.~\ref{dark} shows example results on sunny$\to$night. 
 We randomly select one output for each method. It's obvious that our results are much more realistic, diverse with higher quality. If the object area is small, MUNIT~\cite{huang2018multimodal} may fall into mode collapse and brings small artifacts around object area, in contrast, our method can overcome this problem through instance-level reconstruction.
We also visualize the multimodal results in Fig.~\ref{multimodal} with randomly sampled style vectors. It can be observed that the various degrees of darkness are generated across these images.

\noindent{\textbf{Instance Generation.}} The results of generated instances are shown in Fig.~\ref{instance}, our method can generate more diverse objects (columns 1, 2, 6), more details (columns 5, 6, 7) with even the reflection (column 7). MUNIT sometimes fails to generate desired results if the global style is not suitable for the target object (column 2).

\begin{minipage}[b]{0.35\linewidth}
		\centering
	\includegraphics[height=0.75\textwidth]{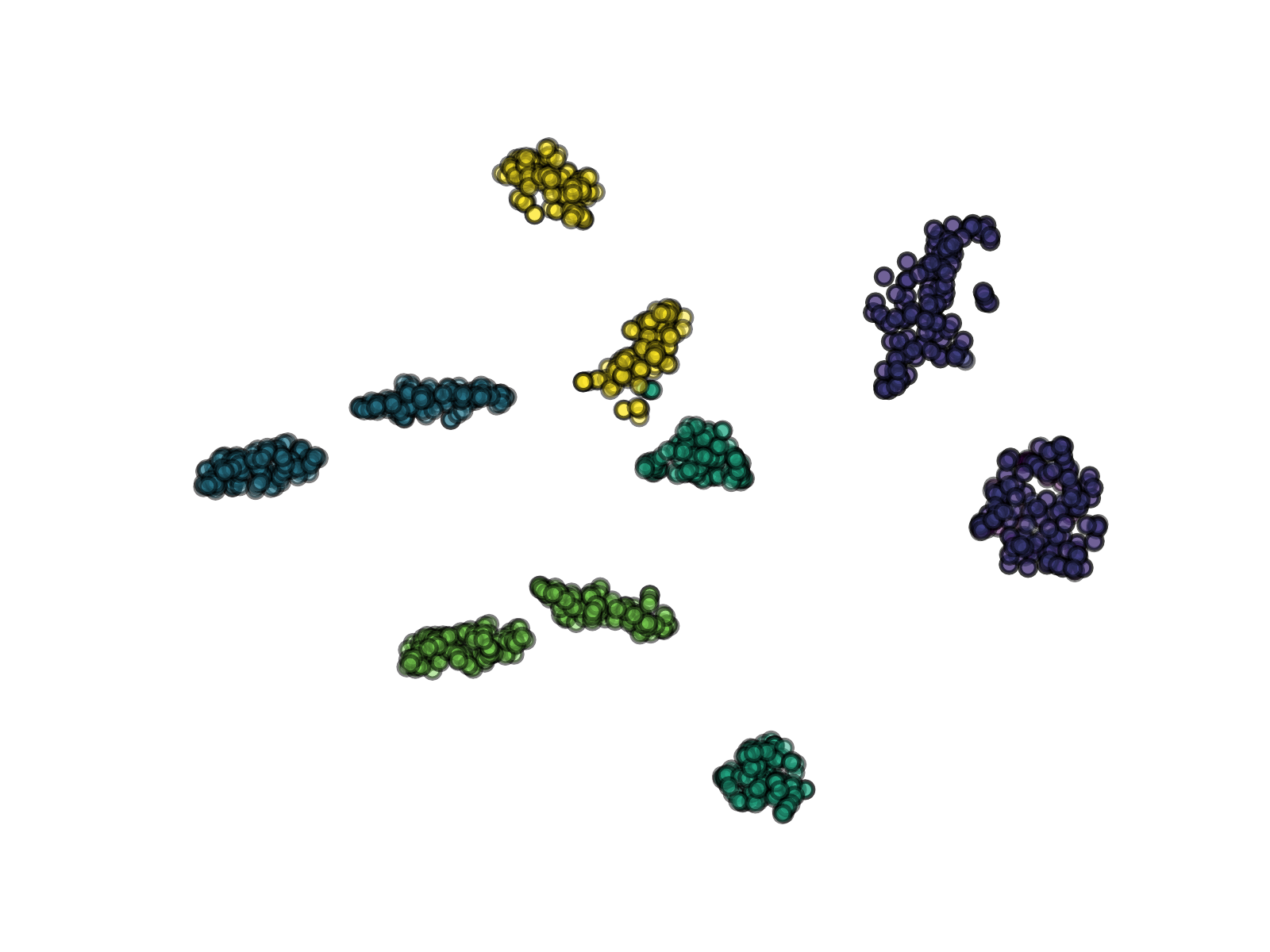}
\end{minipage}
\vspace{0.06in}
\hfill
\begin{minipage}[b]{0.5\linewidth}
	{{\bf Visualization of style distribution by t-SNE~\cite{maaten2008visualizing}.} The groups with the same color are paired object and global styles of same domain.} 
	\label{tsne}
\end{minipage}

\noindent{\textbf{{Comparison of Local (Object) and Global Style Code Distributions.}}
To further verify our assumption that the object and global styles are distinguishable enough to disentangle, we visualize the embedded style vectors from our w/ D$_s$ model. The visualization is plotted by t-SNE tool~\cite{maaten2008visualizing}. We randomly sample 100 images and objects in the test set of each domain, results are shown in Fig.~\ref{tsne}. The same color groups represent the paired global images and objects in the same domain. We can observe that the style vectors of same domain global and object images are grouped and separate with a remarkable margin, meanwhile, they are neighboring in the embedded space. This is reasonable and demonstrates the effectiveness of our learning process.
\section{Conclusion}
In this paper, we have presented a framework for instance-aware I2I translation with unpaired training data. Extensive qualitative and quantitative results demonstrate that the proposed method can capture the details of objects and produce realistic and diverse images. Meanwhile, we also built up a large scale dataset with bounding box annotation for the instance-level I2I translation problem.

\noindent{\textbf{{Acknowledgements}}
Xiangyang Xue was supported in part by NSFC under Grant (No.61572138 \& No.U1611461) and STCSM Project under Grant No.16JC1420400.

{\small
\bibliographystyle{ieee_fullname}
\bibliography{INIT_camera_ready}
}

\end{document}